\documentclass[final]{nesy2026} 

\usepackage{longtable}

\usepackage{booktabs}
\usepackage[load-configurations=version-1]{siunitx} 

\usepackage{comment}
\usepackage{multirow}
\usepackage{wrapfig}

\theorembodyfont{\upshape}
\theoremheaderfont{\scshape}
\theorempostheader{:}
\theoremsep{\newline}

\title[Soft Prototypical Concepts]{Soft Symbol Grounding for Prototypical Concepts}

\author{\Name{Marcos Galván-López} \Email{mgalvanl1600@alumno.ipn.mx}\\
\addr Centro de Investigación en Computación, Instituto Politécnico Nacional. \\
\Name{Nijesh Upreti} \Email{n.upreti@ed.ac.uk}\\
\addr School of Informatics, University of Edinburgh. \\
\Name{Hiram Calvo} \Email{hcalvo@cic.ipn.mx}\\
\addr Centro de Investigación en Computación, Instituto Politécnico Nacional. \\
\Name{Carlos Aguilar-Ibáñez} \Email{carlosaguilari@cic.ipn.mx}\\
\addr Centro de Investigación en Computación, Instituto Politécnico Nacional. \\
\Name{Vaishak Belle} \Email{vbelle@ed.ac.uk}\\
\addr School of Informatics, University of Edinburgh.\\
  }

\begin{document}
\emergencystretch=3em

\maketitle

\begin{abstract}
Neuro-symbolic models are usually trained with supervision only on final labels, leaving the intermediate concepts unobserved. Since many concept assignments are consistent with a given label, training can predict labels correctly while recovering the wrong concepts, a failure known as a reasoning shortcut. Prototypical networks reduce shortcuts by anchoring each concept to a few labeled examples, but existing methods still couple perception and reasoning through a hand-crafted, task-specific differentiable loss that must be redesigned for every task. We introduce \textbf{Soft-PNet}, which removes this loss: it reframes concept grounding as a Metropolis walk over a precomputed cache of feasible symbolic solutions, guided by a prototype distribution built from a single labeled anchor per concept, and trains against one KL objective between the prototype-weighted cache and the network's concept predictions. The objective is identical across tasks and remains applicable when the solution space cannot be enumerated. On \texttt{MNIST-EvenOdd}, Visual Sudoku, and \texttt{Kand-Logic} under scarce supervision, Soft-PNet matches loss-engineered prototypical networks at the concept and label levels and recovers concepts that soft-grounding baselines miss, with no loss engineering and lower training time.

\end{abstract}

\section{Introduction}

Neuro-symbolic (NeSy) AI pursues the combination of neural representational power with the logical rigor of symbolic reasoning \citep{zhou_abductive_2019, sarker_neuro-symbolic_2022, garcez_neural-symbolic_2019}, promising models that are data-efficient and compliant by design with prior knowledge. The canonical NeSy setting is weakly supervised: the model observes raw inputs and final labels, but the intermediate symbolic concepts that connect them remain latent. When the network exploits coincidental patterns in the data to satisfy the symbolic constraints without recovering the true concept semantics, it takes a \emph{reasoning shortcut} \citep{andolfi_right_2025, marconato_not_2023, marconato_bears_2024}.

The difficulty is that a label is consistent with many concept assignments: the feasible set $\mathcal{S}(\mathbf{x}, \mathbf{y})$ of configurations satisfying the background knowledge is typically large, and label supervision alone cannot tell its semantically correct element from the rest. Two lines of work address this. Probabilistic soft grounding explores $\mathcal{S}(\mathbf{x}, \mathbf{y})$ with a projection-based Markov chain \citep{li_softened_2024}, but the chain is steered only by the network's own predictions, which at initialization carry no perceptual information, so reasoning shortcuts reappear within the grounding. Prototypical networks instead inject a perceptual signal, anchoring each concept to a single labeled example and provably reducing shortcuts \citep{andolfi_right_2025, martone_prototypical_2022}, but they still drive learning through a hand-crafted task-specific NeSy loss that must be re-derived for every task and scales poorly with symbolic complexity \citep{badreddine_logic_2022, manhaeve_deepproblog_2018, ahmed_semantic_2022}.

We introduce \textbf{Soft-PNet},\footnote{Code and instructions to reproduce all experiments are available at 
} which unifies these two ideas and removes the loss entirely. The intuition is simple: a single labeled image of each concept already says what that concept looks like, and this is the signal the feasible-set search lacks. In \texttt{MNIST-EvenOdd}, for instance, a label fixes the parity of a digit sum but not the digits, and only the perceptually faithful assignment aligns with the anchor prototypes. Soft-PNet precomputes a cache of feasible solutions with any solver or generator, scores and refines it with a prototype-guided Metropolis update, and defines the training loss as a Kullback-Leibler (KL) divergence between the prototype-weighted cache and the network's predictions. The prototype scores supply the perceptual signal missing from \citet{li_softened_2024}, while cache membership encodes the constraints implicitly, replacing the task-specific loss of \citet{andolfi_right_2025} with one generic objective identical across tasks.

We evaluate \textbf{Soft-PNet} on \texttt{MNIST-EvenOdd}, Visual Sudoku Classification, and \texttt{Kand-Logic} under scarce supervision, with one labeled anchor per concept. Against the soft-grounding baseline of \citet{li_softened_2024}, which explores the feasible set without perceptual guidance, Soft-PNet raises digit-level accuracy from $27\%$ to $96\%$ on \texttt{MNIST-EvenOdd} and from $32\%$ to $99\%$ on Visual Sudoku. Against the loss-engineered prototypical network of \citet{andolfi_right_2025}, it matches or exceeds concept-level accuracy and stays within one standard deviation at the label level, while removing the task-specific loss entirely and reducing per-epoch training time by up to $50\%$. These results show that perceptual anchoring and cache-based exploration are mutually necessary: neither suffices alone, but together they ground concepts correctly with no loss engineering.

\section{Related Work}

\textbf{NeSy grounding and reasoning shortcuts.} A broad range of NeSy methods couple perception and reasoning by softening logical semantics: probabilistic \citep{manhaeve_deepproblog_2018, choi_probabilistic_2020} or fuzzy \citep{badreddine_logic_2022} relaxations, compilation of background knowledge into a differentiable loss \citep{ahmed_semantic_2022, wang_satnet_2019}, abductive inference \citep{zhou_abductive_2019}, and closed-loop program execution over learned concepts \citep{li_closed_2020, mao_neuro-symbolic_2019}. When the intermediate concepts are latent, such systems may take reasoning shortcuts that satisfy the label constraint without recovering the true concepts \citep{marconato_not_2023, marconato_neuro-symbolic_2023, shao_right_2021, marconato_bears_2024}. Concept Bottleneck Models \citep{koh_concept_2020} and their extensions \citep{espinosa_zarlenga_concept_2022, debot_interpretable_2024, kim_interpretability_2018} interpose an explicit concept layer but assume a fixed vocabulary and remain prone to leakage under scarce supervision. We instead operate in the extreme low-annotation regime and model the combinatorial feasible set explicitly.

\textbf{Soft and prototypical grounding.} Closest to our work, \citet{li_softened_2024} model the feasible set as a Boltzmann distribution and sample it by a projection-based Markov chain Monte Carlo, but their single-candidate chain is steered only by the network's own predictions, which carry no perceptual signal at initialization. Prototypical networks \citep{snell_prototypical_2017} supply that signal, anchoring each concept to one labeled example; their neuro-symbolic adaptations \citep{martone_prototypical_2022, andolfi_right_2025} provably reduce shortcuts under mild separability, yet still require a hand-crafted task-specific NeSy loss re-derived per task. We combine the two: a prototype distribution scores a precomputed cache of feasible solutions, providing the signal absent from \citet{li_softened_2024} while replacing the task-specific loss of \citet{andolfi_right_2025} with a single generic KL objective. \appendixref{apd:related} gives an extended discussion, including how our cache-based procedure relates to the shortcut analysis of \citet{marconato_bears_2024}.

\section{Problem Statement}

\textbf{Notation}:
Scalar constants are denoted by lower-case letters $x$, while random variables are denoted by upper-case letters $X$. Ordered sets of constants and random variables are represented in boldface as $\mathbf{x}$ and $\mathbf{X}$, respectively. We use the notation $[n]$ to denote the set $\{1, \dots, n\}$, and the prior knowledge is denoted by $\textbf{K}$.

We consider a NeSy system that infers labels $\mathbf{Y}$ by reasoning over the background knowledge $\mathbf{K}$ and a set of $k$ discrete concepts extracted from a sub-symbolic input $\mathbf{X}$. We posit a ground-truth generative process $p^*(\mathbf{X}, \mathbf{G}, \mathbf{Y})$ in which $\mathbf{G} \in \mathcal{C} = [h_1] \times \cdots \times [h_k]$ are the true latent concepts, and the background knowledge induces a reasoning function $f_\mathbf{K}: \mathcal{C} \rightarrow \mathcal{Y}$ with $\mathbf{y} = f_\mathbf{K}(\mathbf{g})$ for every assignment $\mathbf{g} \in \mathcal{C}$. A NeSy predictor produces a concept distribution $p_\theta(\mathbf{C} \mid \mathbf{X})$ and the induced label distribution $p_\theta(\mathbf{Y} \mid \mathbf{X}; \mathbf{K})$. Our objective is to learn the parameters $\theta$ so that $p_\theta(\mathbf{C} \mid \mathbf{x})$ matches the true concept posterior $p^*(\mathbf{G} \mid \mathbf{x})$ for all $\mathbf{x} \in \mathcal{X}$.

Given a dataset $\mathcal{D} = \{(\mathbf{x}_i, \mathbf{y}_i)\}^n_{i=1}$, the set of concept assignments globally consistent with an observed pair $(\mathbf{x}, \mathbf{y})$ under the reasoning function is

$$
\mathcal{S}(\mathbf{x}, \mathbf{y}) := \{ \mathbf{c} \in \mathcal{C} \mid f_\mathbf{K} (\mathbf{c}) = \mathbf{y}\}
$$

This is the feasible set for $(\mathbf{x}, \mathbf{y})$, where every element of $\mathcal{S}(\mathbf{x}, \mathbf{y})$ is a concept assignment that is consistent with the observed label under the background knowledge. The label likelihood marginalizes the concept distribution over this set, $p_\theta(\mathbf{y} \mid \mathbf{x}; \mathbf{K}) = \sum_{\mathbf{c} \in \mathcal{S}(\mathbf{x}, \mathbf{y})} p_\theta(\mathbf{c} \mid \mathbf{x})$, so it is unchanged by how probability mass is distributed within $\mathcal{S}(\mathbf{x}, \mathbf{y})$. Since $|\mathcal{S}(\mathbf{x}, \mathbf{y})| > 1$ in general, maximizing it cannot identify which element of $\mathcal{S}(\mathbf{x}, \mathbf{y})$ corresponds to the true underlying concepts. Any assignment in the feasible set is an equally valid solution from the perspective of the loss, making the training objective blind to semantic correctness at the concept level.

\subsection{Reasoning Shortcuts}

The ambiguity of $\mathcal{S}(\mathbf{x}, \mathbf{y})$ gives rise to the reasoning shortcut problem. Let $\mathcal{D} = \{(\mathbf{x}_i, \mathbf{y}_i)\}^n_{i=1}$ be a finite dataset. We say that a NeSy predictor $p_\theta$ takes a \emph{reasoning shortcut} if it achieves maximum log-likelihood on $\mathcal{D}$ while failing to recover the true concept distributions, that is

\begin{align}
\mathcal{L}(p_\theta, \mathcal{D}, \mathbf{K}) = \max_{\theta' \in \Theta} \mathcal{L}(p_{\theta'}, \mathcal{D}, \mathbf{K}) \quad \text{and} \quad p_\theta(\mathbf{C} \mid \mathbf{X}) \neq p^*(\mathbf{G} \mid \mathbf{X}), \notag
\end{align}

where $\mathcal{L}(p_\theta, \mathcal{D}, \mathbf{K}) = \frac{1}{|\mathcal{D}|} \sum_{(\mathbf{x}, \mathbf{y}) \in \mathcal{D}} \log p_\theta(\mathbf{y} \mid \mathbf{x}; \mathbf{K})$. Every reasoning shortcut corresponds to a deterministic optimum of $\mathcal{L}$  \citep{andolfi_right_2025, marconato_not_2023}, meaning that standard gradient-based training provides no mechanism to escape them. The model has found a concept assignment inside $\mathcal{S}(\mathbf{x}, \mathbf{y})$ that is semantically wrong but logically indistinguishable from the correct one under label supervision alone.

\paragraph{Example.} In \texttt{MNIST-EvenOdd} a training pair is two digit images labeled only by their sum. For the label $\mathbf{y}=10$, the feasible set $\mathcal{S}(\mathbf{x},\mathbf{y}) = \{(1,9),(2,8),(3,7),\dots,(8,2),(9,1)\}$ contains nine assignments consistent with the label, and supervision cannot prefer any one of them. A network can therefore map every pair summing to $10$ to $(8,2)$, satisfying the constraint without ever learning to read digits. A single labeled image per digit fixes what each concept looks like, so only the assignment whose images match these anchors is perceptually correct, breaking the tie the label leaves open.

\section{Soft-PNet}

Soft grounding \citep{li_softened_2024} samples a tempered Boltzmann distribution over $\mathcal{S}(\mathbf{x}, \mathbf{y})$ with a projection-based Metropolis walk, sharpened toward a deterministic grounding (\appendixref{apd:background}). Two problems remain: a single candidate is kept per data point, so a poor initialization is rarely escaped, and the walk is steered only by $p_\theta$, which at initialization carries no perceptual information. Soft-PNet resolves both with a $K$-candidate solution cache scored by a prototype distribution.

\begin{figure}[t]
    \centering
    \includegraphics[width=\linewidth]{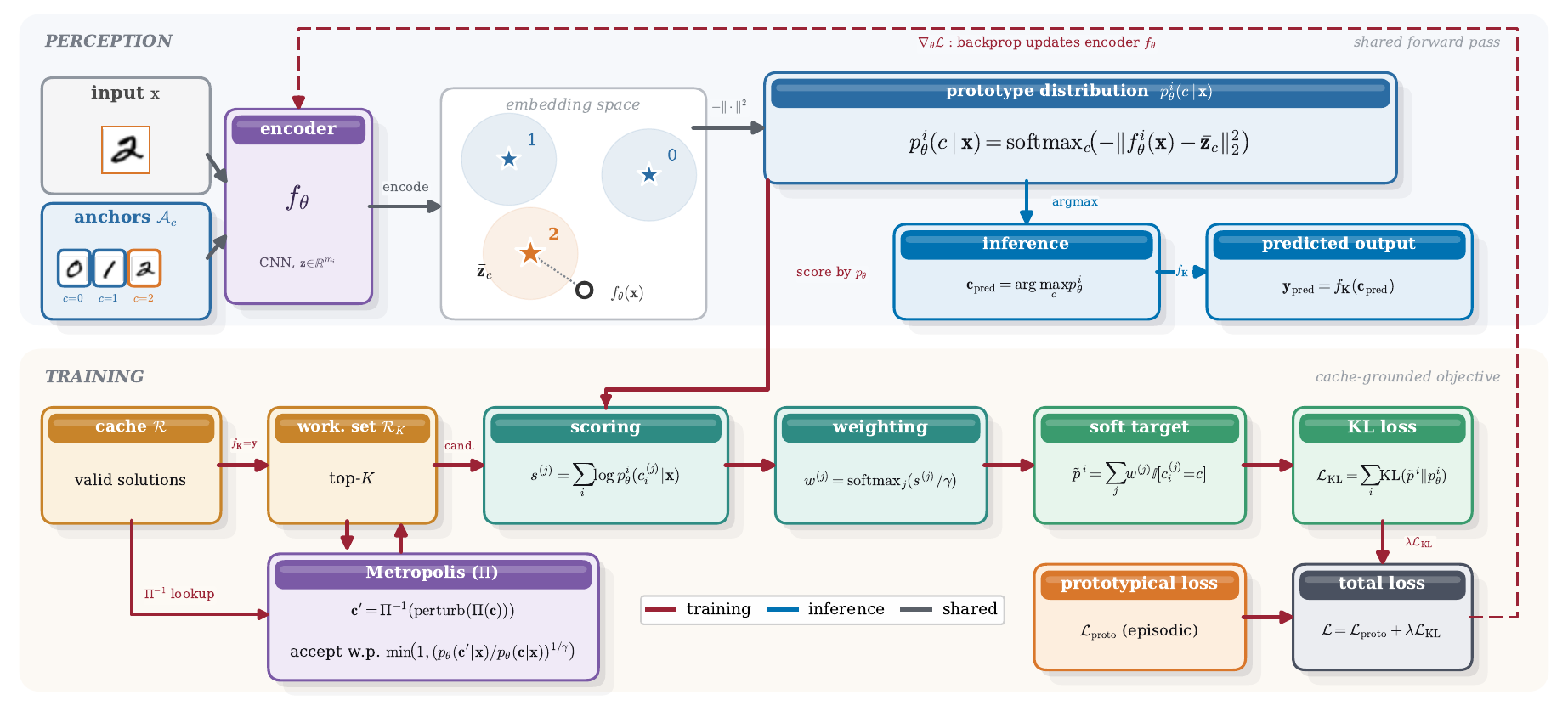}
    \caption{Overview of Soft-PNet. A single labeled anchor per concept builds the prototypes $\bar{\mathbf{z}}_c$ that define the prototype distribution $p_\theta^i$ (top). \textbf{Training} (red): the working set $\mathcal{R}_K$ is drawn from the precomputed cache $\mathcal{R}$ of valid solutions by keeping the top-$K$ feasible candidates ($f_\mathbf{K}(\mathbf{c}) = \mathbf{y}$), refined by the projection-based Metropolis walk, and scored against $p_\theta$; the scores are tempered into weights $w^{(j)}$ whose per-concept marginal forms the soft target of the KL loss, and the objective adds the episodic prototypical loss, with gradients of the total objective updating the encoder $f_\theta$. \textbf{Inference} (blue): the predicted concepts $\mathbf{c}_{\mathrm{pred}}$ are the per-group argmax of $p_\theta^i$, and the label $\mathbf{y}_{\mathrm{pred}} = f_\mathbf{K}(\mathbf{c}_{\mathrm{pred}})$ follows from the reasoning function. Gray arrows are shared between the two.}
    \label{fig:pipeline}
\end{figure}

\subsection{Solution Cache}

We maintain a global cache of precomputed feasible solutions and select a working set of $K$ candidates per training step, so the chain always has alternatives to a poorly initialized grounding.

Before training begins, we generate a representative subset of the global solution space and store it as a tensorized lookup table

$$\mathcal{R} = \{ \mathbf{c}^{(1)}, \dots, \mathbf{c}^{(M_0)}\}$$

using any available solver or combinatorial generator. This precomputation is done once and decouples solution generation from the training loop. The tensorized structure of $\mathcal{R}$ allows scoring and Metropolis updates to be computed in parallel across the entire working set at each step.

At the first time a training pair $(\mathbf{x}, \mathbf{y})$ is encountered, its working set $\mathcal{R}_K$ is initialized by filtering $\mathcal{R}$ to feasible solutions, drawing $M = K \cdot N$ candidates, scoring them against the current network output, and retaining the top-$K$, as described in \algorithmref{alg:init_cache}. On subsequent encounters, $\mathcal{R}_K$ is refined via the projection-based Metropolis update in \algorithmref{alg:metro_cache}.
Maintaining $K$ candidates rather than one reduces the variance of the gradient estimate, but it does not resolve the semantic problem: both the top-$K$ scoring (\algorithmref{alg:init_cache}) and the acceptance ratio (\algorithmref{alg:metro_cache}) are driven entirely by $p_\theta(\mathbf{c} \mid \mathbf{x})$, which at initialization carries no information about true concept identity, leaving the method vulnerable to reasoning shortcuts.

\subsection{Projection Operator}

The Metropolis update in \algorithmref{alg:metro_cache} relies on a projection operator $\Pi: \mathcal{C} \rightarrow \Omega$ that maps a feasible state to a lower-dimensional space where connectivity is improved, applies a perturbation, and recovers a new feasible state via the inverse projection $\Pi^{-1}$.

For example, in \texttt{MNIST-EvenOdd} a state is a digit pair $(d_1, d_2)$ with $d_1 + d_2 = n$; the projection reduces it to the sum, and the inverse projection samples a different feasible pair with that sum from the cache. In general $\Pi$ drops a subset of coordinates, the perturbation modifies the rest, and $\Pi^{-1}$ reconstructs a feasible completion by cache lookup; for Visual Sudoku it swaps two cells within a diagonal sub-block and for \texttt{Kand-Logic} it switches one object attribute (\appendixref{apd:projection}).

\subsection{Prototype-Anchored Cache and KL Loss}

To address this, we replace the perceptually uninformed scoring function with one grounded in the observed data. We introduce an anchor set $\mathcal{A}_c$ for each concept $c \in [h_i]$, consisting of a single labeled example per concept and a set of data-augmented samples derived from it. From these anchors, we construct a prototype vector $\bar{\mathbf{z}}_c$ as the centroid of the anchor embeddings, $\bar{\mathbf{z}}_c = \frac{1}{|\mathcal{A}_c|} \sum_{\mathbf{x}_a \in \mathcal{A}_c} f_\theta^i(\mathbf{x}_a)$, in the latent space of a prototypical network $f_\theta^i : \mathcal{X} \rightarrow \mathbb{R}^{m_i}$.

Given a new input $\mathbf{x}$, the concept distribution is defined as a softmax over negative squared distances to the prototype vectors

\begin{align}
    p_\theta^i(c \mid \mathbf{x}) = \frac{\exp(-\| f_\theta^i(\mathbf{x}) - 
    \bar{\mathbf{z}}_c \|_2^2)}{\sum_{c' \in [h_i]} \exp(-\| f_\theta^i(\mathbf{x}) 
    - \bar{\mathbf{z}}_{c'} \|_2^2)}, \quad c \in [h_i],
\end{align}

The factorized joint over all concept groups is $p_\theta(\mathbf{c} \mid \mathbf{x}) := \prod_{i=1}^k p_\theta^i(c_i \mid \mathbf{x})$ for $\mathbf{c} = (c_1, \dots, c_k) \in \mathcal{C}$. This distribution encodes which concept assignments are geometrically consistent with the observed input relative to the labeled anchors, replacing the uninformed classifier output with a perceptually grounded scoring function from initialization. Substituting it for $p_\theta$ in the acceptance ratio of \algorithmref{alg:metro_cache} and the top-$K$ scoring of \algorithmref{alg:init_cache} biases the chain toward candidates that are both feasible and perceptually coherent with the anchor embeddings.

\paragraph{KL Loss.} Given the working set $\mathcal{R}_K(\mathbf{x}, \mathbf{y})$, we score each candidate by its joint log-probability $s^{(j)} = \log p_\theta(\mathbf{c}^{(j)} \mid \mathbf{x}) = \sum_{i=1}^k \log p_\theta^i(c_i^{(j)} \mid \mathbf{x})$ and weight it with a temperature-scaled softmax over the $K$ scores, $w^{(j)} = \exp(s^{(j)} / \gamma) / \sum_{l=1}^K \exp(s^{(l)} / \gamma)$. Rather than treating the $K$ candidates as a distribution over full concept assignments directly, we marginalize over the cache to obtain a soft target distribution over each concept independently. For each concept position $i \in [k]$ and each concept value $c \in [h_i]$, the soft target is

$$\tilde{p}^i(c \mid \mathbf{x}) = \sum_{j=1}^K w^{(j)} \cdot \mathbb{I}\!\left[c_i^{(j)} = c\right].$$

This gives a per-concept marginal distribution that concentrates mass on values that appear frequently among the high-scoring candidates. The NeSy loss is then computed as the KL divergence between these soft targets and the network's concept predictions

\begin{align}\label{eq:kl_loss}
    \mathcal{L}_{\mathrm{KL}}(\theta) = \sum_{(\mathbf{x}, \mathbf{y}) \in 
    \mathcal{D}} \sum_{i=1}^k \sum_{c \in [h_i]} \tilde{p}^i(c \mid \mathbf{x}) 
    \log \frac{\tilde{p}^i(c \mid \mathbf{x})}{p_\theta^i(c \mid \mathbf{x})}.
\end{align}

The soft targets $\tilde{p}^i$ are treated as fixed when differentiating, so minimizing \equationref{eq:kl_loss} reduces to a cross-entropy that pulls the network's concept posterior toward the cache-induced marginals, assigning high probability to concept values consistent with the highest-scoring feasible solutions. The symbolic constraints are encoded implicitly through cache membership, and no task-specific differentiable NeSy loss needs to be designed. The same objective applies across tasks, with the cache $\mathcal{R}$, projection $\Pi$, and target temperature varying between them.

\paragraph{Prototypical Loss.} The prototypical loss $\mathcal{L}_{\mathrm{proto}}$ is the standard episodic prototypical-networks loss \citep{snell_prototypical_2017} over the augmented anchor set. Each episode partitions the anchors $\mathcal{A}_c$ into a support set $\mathbf{S}_c$ and a query set $\mathbf{Q}_c$, forms each class prototype $\boldsymbol{\mu}_c$ as the centroid of its support embeddings, and minimizes the query negative log-likelihood $\mathcal{L}_{\mathrm{proto}} = -\frac{1}{|\mathbf{Q}|} \sum_{(\mathbf{x}_q, c_q) \in \mathbf{Q}} \log \mathrm{softmax}_{c_q}\!\big(-\| f_\theta^i(\mathbf{x}_q) - \boldsymbol{\mu}_{c} \|_2^2\big)$, the softmax taken over the squared distances from $f_\theta^i(\mathbf{x}_q)$ to all prototypes $\boldsymbol{\mu}_c$. This keeps the embedding space clustered around the anchor prototypes; it is applied as a warmup before the cache loss is introduced and at each minibatch step thereafter.

The full training objective combines the KL loss with an episodic prototypical loss $\mathcal{L}_{\mathrm{proto}}$ over the labeled anchors, keeping the embedding space structured throughout training:

\begin{align}\label{eq:total_loss}
    \mathcal{L}(\theta) = \mathcal{L}_{\mathrm{proto}}(\theta) +
    \lambda \cdot \mathcal{L}_{\mathrm{KL}}(\theta),
\end{align}

where $\lambda$ weights the cache loss; it is fixed per task and, on \texttt{Kand-Logic}, ramped up from zero over the first epochs so the embedding space stabilizes before the cache signal is introduced \citep{andolfi_right_2025, martone_prototypical_2022}.

An overview of the complete training procedure is given in \figureref{fig:pipeline} and \algorithmref{alg:softpnet_training}; it builds each working set $\mathcal{R}_K$ with \algorithmref{alg:init_cache} and refines it with the Metropolis update \algorithmref{alg:metro_cache}, both detailed in \appendixref{apd:training}.

\begin{algorithm2e}[t]
\caption{Soft-PNet Training}
\label{alg:softpnet_training}
\KwIn{Dataset $\mathcal{D}$, anchors $\mathcal{A}_c$, cache $\mathcal{R}$, epochs $E$, warmup $E_w$, schedules $\gamma(e), \lambda(e)$, integers $K, N$}
\KwOut{Parameters $\theta$}
Augment anchors to $\hat{\mathcal{A}}_c$;\quad $\text{initialized}[\cdot] \leftarrow \texttt{false}$\;
\For{$e = 1$ \KwTo $E$}{
    $\gamma \leftarrow \gamma(e)$,\quad $\lambda \leftarrow \lambda(e)$\;
    \For{minibatch $\mathcal{B} \subseteq \mathcal{D}$}{
        Compute prototypes $\bar{\mathbf{z}}_c$ and prototypical loss $\mathcal{L}_{\mathrm{proto}}$ from $\hat{\mathcal{A}}_c$\;
        $\mathcal{L}_{\mathrm{KL}} \leftarrow 0$\;
        \For{$(\mathbf{x}, y, \text{id}) \in \mathcal{B}$}{
            \eIf{$\neg\,\text{initialized}[\text{id}]$}{
                Initialize $\mathcal{R}_K[\text{id}]$ from $(\mathbf{x}, y)$ via \algorithmref{alg:init_cache};\quad $\text{initialized}[\text{id}] \leftarrow \texttt{true}$\;
            }{
                Update $\mathcal{R}_K[\text{id}]$ from $\mathbf{x}$ via \algorithmref{alg:metro_cache}\;
            }
            \lIf{$e > E_w$}{add the KL term of $\mathbf{x}$ from $\mathcal{R}_K[\text{id}]$ to $\mathcal{L}_{\mathrm{KL}}$ (\equationref{eq:kl_loss})}
        }
        $\mathcal{L} \leftarrow \mathcal{L}_{\mathrm{proto}} + \lambda\, \mathcal{L}_{\mathrm{KL}}$;\quad update $\theta$\;
    }
}
\end{algorithm2e}

\section{Experimental Analysis}

We evaluate \textbf{Soft-PNet} on three tasks under scarce supervision, with a single labeled anchor per concept and results averaged over ten seeds. We report concept-level and label-level performance and training time per epoch; the concept-label gap is the primary diagnostic for reasoning shortcuts, since high label accuracy with low concept accuracy signals a spurious grounding.

\textbf{Tasks.} (i) \texttt{MNIST-EvenOdd} \citep{marconato_bears_2024, marconato_neuro-symbolic_2023}: each input is a pair of handwritten digits whose sum is the observed label, restricted to same-parity pairs so the mixed-parity test set is out of distribution. (ii) Visual Sudoku Classification \citep{wang_satnet_2019}: a $4 \times 4$ Sudoku board populated with MNIST digits, labeled as valid or invalid. (iii) \texttt{Kand-Logic} \citep{muller_kandinsky_2021, marconato_bears_2024}: scenes composed of geometric primitives defined by shape and color, labeled according to a logical pattern. This task has an intractable solution space, making it a direct test of the cache-based approach under combinatorial complexity.

\textbf{Baselines.} For \texttt{MNIST-EvenOdd} and Visual Sudoku we compare against four methods: the original Soft Grounding (\textbf{SoftG}) \citep{li_softened_2024}, a $K$-sample extension of Soft Grounding without prototypical anchoring (\textbf{SoftG-K}), Semantic Loss with prototypical networks (\textbf{PNet+SL}) \citep{andolfi_right_2025}, and our method (\textbf{Soft-PNet}). For \texttt{Kand-Logic}, we compare against \textbf{PNet+SL} only, as the intractable solution space makes the SoftG variants inapplicable.

The reproducibility details, neural architectures, and hyperparameter calibration are specified in \appendixref{apd:setup}.

\subsection{MNIST-EvenOdd}

Results are reported in \tableref{tab:results}. Both SoftG and SoftG-K collapse at the concept level, achieving digit accuracy below $28\%$ despite non-trivial sum accuracy. This is a reasoning shortcut: the models satisfy the arithmetic constraint through spurious digit assignments rather than digit recognition. The addition of $K$ candidates without perceptual anchoring (SoftG-K) does not resolve this, confirming that cache diversity alone is insufficient.

\textbf{Soft-PNet} outperforms PNet+SL on digit accuracy while substantially reducing variance, indicating that the prototype-guided cache provides a more stable training signal than the NeSy loss alone in the low-data regime. Sum-level F1 is omitted because class imbalance across possible sums makes it an unreliable label-level indicator, so sum accuracy is reported instead. Training time per epoch is slightly lower than PNet+SL ($60.2$ vs $70.2$ s), as the cache removes the per-step logical loss.

\subsection{Visual Sudoku Classification}

The concept-label gap for SoftG and SoftG-K is large (\tableref{tab:results}): both achieve board accuracy above $74\%$ while digit accuracy remains below $32\%$. The Metropolis chain explores the symmetry group of valid Sudoku boards without any perceptual anchor to break the digit permutation ambiguity, producing models that classify boards correctly through structural pattern matching rather than digit recognition.

\textbf{Soft-PNet} closes this gap, matching PNet+SL at both the concept and label levels while training in $21.5$ s per epoch against $43.4$, a roughly $50\%$ reduction, since the Sudoku constraint is encoded implicitly through the cache rather than a differentiable loss over all row, column, and subgrid pairs.

\subsection{Kand-Logic}

On \texttt{Kand-Logic} (\tableref{tab:results}), the solution space is combinatorially large, making exhaustive enumeration intractable. \textbf{Soft-PNet} operates on a precomputed representative subset of feasible scenes, showing that the method does not require complete coverage of $\mathcal{S}(\mathbf{x}, \mathbf{y})$.

\textbf{Soft-PNet} matches PNet+SL to within one standard deviation at both levels while training faster ($76.5$ vs $89.8$ s per epoch), confirming that the prototype-guided cache achieves competitive grounding without a hand-crafted loss even when the solution space must be approximated.

\begin{table}[t]
\floatconts
    {tab:results}
    {\caption{Results on all three tasks (mean $\pm$ std over ten seeds). Concept-level metrics refer to digit (\texttt{MNIST-EvenOdd}, Visual Sudoku) or primitive (\texttt{Kand-Logic}) recognition; label-level metrics refer to the sum, board validity, or scene class. \texttt{MNIST-EvenOdd} label F1 is omitted due to class imbalance across sums. Training sizes are $6{,}720$ pairs, $300$ boards, and $3{,}000$ scenes, following \citet{andolfi_right_2025, li_softened_2024}.}}
    {\small\setlength{\tabcolsep}{4pt}
    \begin{tabular}{llcccc}
    \hline
    \textbf{Task} & \textbf{Method} & \textbf{Concept Acc} & \textbf{Concept F1} & \textbf{Label Acc} & \textbf{Label F1} \\ \hline
    \multirow{4}{*}{\texttt{MNIST-EvenOdd}}
      & SoftG     & $27.33 \pm 11.46$ & $17.78 \pm 8.60$ & $5.11 \pm 4.71$ & n/a \\
      & SoftG-K   & $26.61 \pm 10.64$ & $15.55 \pm 7.19$ & $3.96 \pm 4.79$ & n/a \\
      & PNet+SL   & $85.40 \pm 17.18$ & $83.33 \pm 19.72$ & $73.76 \pm 29.83$ & n/a \\
      & \textbf{Soft-PNet} & $\mathbf{96.44 \pm 6.28}$ & $\mathbf{95.86 \pm 7.98}$ & $\mathbf{92.99 \pm 12.22}$ & n/a \\ \hline
    \multirow{4}{*}{Visual Sudoku}
      & SoftG     & $31.78 \pm 16.04$ & $31.52 \pm 16.07$ & $75.76 \pm 9.65$ & $65.65 \pm 21.26$ \\
      & SoftG-K   & $27.05 \pm 20.31$ & $27.21 \pm 20.23$ & $74.50 \pm 9.48$ & $63.32 \pm 21.70$ \\
      & PNet+SL   & $98.79 \pm 0.33$ & $98.79 \pm 0.33$ & $91.21 \pm 2.30$ & $90.31 \pm 2.78$ \\
      & \textbf{Soft-PNet} & $\mathbf{98.60 \pm 0.62}$ & $\mathbf{98.60 \pm 0.61}$ & $89.82 \pm 4.00$ & $88.48 \pm 5.18$ \\ \hline
    \multirow{2}{*}{\texttt{Kand-Logic}}
      & PNet+SL   & $91.45 \pm 2.28$ & $81.78 \pm 5.42$ & $83.84 \pm 3.70$ & $84.02 \pm 3.64$ \\
      & \textbf{Soft-PNet} & $90.26 \pm 2.16$ & $79.18 \pm 3.49$ & $81.60 \pm 7.38$ & $81.81 \pm 7.65$ \\ \hline
    \end{tabular}}
\end{table}

\section{Discussion}

The advantage is largest on \texttt{MNIST-EvenOdd}, whose test set is out of distribution (mixed-parity pairs absent from training): the anchors give a perceptual signal that generalizes to these unseen pairs, so Soft-PNet improves accuracy and sharply reduces variance, while the large standard deviations of PNet+SL reflect seeds that collapse into shortcuts. On Visual Sudoku and \texttt{Kand-Logic}, where the test distribution matches training and both methods are near ceiling, Soft-PNet matches PNet+SL and its benefit is the removal of loss engineering and lower training cost.

The ablation in \tableref{tab:ablation}, detailed in \appendixref{apd:ablation}, identifies which component is responsible for correct grounding on Visual Sudoku. Removing the prototype anchor is the most damaging: board F1 falls from $88.5\%$ to $32.0\%$, indicating that the perceptual signal carried by the anchors, rather than the search over the feasible set, is what ties the grounding to the true concepts. Collapsing the working set to a single candidate is also harmful, since one sample cannot estimate the per-concept soft target reliably and board F1 drops to near zero, whereas removing only the projection walk leaves performance essentially unchanged. Prototype-guided scoring is therefore the primary mechanism, the candidate population stabilizes its soft target, and the projection adds exploration.

These results also separate the two prior approaches. Exploring the feasible set without a perceptual signal, as in SoftG and SoftG-K, concentrates probability on structurally valid but semantically arbitrary solutions and reinforces the shortcut rather than resolving it. Anchoring concepts to prototypes, as in PNet+SL, supplies that signal but couples learning to a hand-crafted differentiable loss. Soft-PNet keeps the prototype signal while encoding the symbolic constraints through cache membership, so the training objective reduces to a single KL divergence that is identical across tasks. It matches PNet+SL at both the concept and label levels without this per-task loss, and at lower cost: per-epoch training time is reduced by $14\%$ on \texttt{MNIST-EvenOdd} and $50\%$ on Visual Sudoku, because the objective retains the same structure regardless of how many symbolic operations a task involves.

Two consequences follow. First, a fixed objective removes the main engineering cost of moving these methods to a new task, and because evaluating it does not scale with symbolic complexity, the efficiency gap with hand-crafted losses is expected to grow on larger problems such as multi-digit addition. Second, the \texttt{Kand-Logic} results show that the cache need not enumerate $\mathcal{S}(\mathbf{x}, \mathbf{y})$: a representative subset diverse enough to expose the prototype distribution to a broad range of feasible configurations suffices, which lets the method operate when the feasible set is intractable.

\begin{table}[hbtp]
\floatconts
    {tab:ablation}
    {\caption{Ablation on Visual Sudoku. Each variant removes one component of Soft-PNet: the prototype anchor (Anchor), the $K$-candidate population (Pop.), or the projection walk (Proj.).}}
    {\footnotesize\setlength{\tabcolsep}{3pt}
    \begin{tabular}{lccccccc}
    \hline
    \textbf{Variant} & \textbf{Anchor} & \textbf{Pop.} & \textbf{Proj.} & \textbf{Digit Acc} & \textbf{Digit F1} & \textbf{Board Acc} & \textbf{Board F1} \\ \hline
    Soft-PNet (full)  & \checkmark & \checkmark & \checkmark & $98.60 \pm 0.62$ & $98.60 \pm 0.61$ & $89.82 \pm 4.00$ & $88.48 \pm 5.18$ \\
    Abl-A (no anchor) & $\times$ & \checkmark & \checkmark & $89.97 \pm 3.72$ & $89.86 \pm 3.78$ & $59.88 \pm 5.89$ & $32.03 \pm 16.34$ \\
    Abl-B (no pop.)   & \checkmark & $\times$ & \checkmark & $72.99 \pm 10.92$ & $72.31 \pm 10.73$ & $50.64 \pm 0.51$ & $2.70 \pm 2.10$ \\
    Abl-C (no proj.)  & \checkmark & \checkmark & $\times$ & $98.46 \pm 0.66$ & $98.46 \pm 0.66$ & $88.87 \pm 4.49$ & $87.25 \pm 5.87$ \\ \hline
    \end{tabular}}
\end{table}

\section{Conclusion}

We introduced \textbf{Soft-PNet}, which replaces hand-crafted task-specific NeSy losses with a prototype-guided solution cache and a KL objective. With one labeled anchor per concept, it matches loss-engineered prototypical networks across \texttt{MNIST-EvenOdd}, Visual Sudoku, and \texttt{Kand-Logic} while recovering concepts that soft-grounding baselines miss. Prototype anchoring and cache-based exploration are mutually necessary.

\textbf{Limitations.} Soft-PNet depends on a global precomputed cache $\mathcal{R}$: when the solution space is truly intractable and hard to sample representatively, cache quality directly bounds grounding quality, and highly irregular or sparsely connected feasible sets may need more sophisticated construction than the combinatorial generators used here.

\textbf{Future work.} Natural extensions include concept-level attention over cache candidates and scaling to multi-digit \texttt{MNIST-Addition} and real-world benchmarks such as BDD-OIA \citep{xu_explainable_2020}.

\acks{This work was done with the support of (i) Secretaría de Investigación y Posgrado (SIP) of the Instituto Politécnico Nacional (IPN), under the project number 20260296. (ii) The Secretaría de Ciencia, Humanidades, Tecnología e Innovación (SECIHTI) of the Mexican Government. (iii) The School of Informatics of the University of Edinburgh.}

\bibliography{nesy2026-final}
\newpage

\appendix

\section{Extended Related Work}\label{apd:related}

The main text summarizes related work briefly for space. Here we expand on the lines of work most directly connected to Soft-PNet and state how our method relates to each.

\paragraph{Reasoning shortcuts.} A reasoning shortcut is a parameterization that maximizes the label likelihood while mapping inputs to the wrong concepts \citep{marconato_not_2023, marconato_neuro-symbolic_2023}. \citet{marconato_not_2023} show that these solutions are optima of the standard NeSy objective rather than optimization failures, so they cannot be removed by better training alone, and \citet{marconato_bears_2024} give a procedure for detecting when a trained model is liable to them. Mitigations in the literature include concept-level supervision, architectural priors, and reconstruction objectives \citep{marconato_neuro-symbolic_2023, shao_right_2021}. Soft-PNet uses the minimal form of concept supervision, a single labeled example per concept, to break the tie that label supervision leaves open, rather than to supervise the concept layer densely.

\paragraph{Softened symbol grounding.} The closest technical predecessor is the softened grounding of \citet{li_softened_2024}, described in detail in \appendixref{apd:background}. It represents the feasible groundings of an example as a tempered Boltzmann distribution and samples it with a projection-based Metropolis walk, annealing the temperature toward a single deterministic grounding. Because the walk is steered only by the network's own predictions and keeps one candidate per example, a grounding that is wrong at initialization is rarely corrected: there is no perceptual signal favoring the true concepts and no diversity to escape a poor start. Soft-PNet keeps the Boltzmann formulation and the projection walk but replaces the single candidate with a $K$-candidate cache and the network-only score with a prototype distribution, so the same sampler is driven by a perceptually grounded signal.

\paragraph{Prototypical and few-shot grounding.} Prototypical networks \citep{snell_prototypical_2017} represent each class by the centroid of a few labeled embeddings and classify by distance in that space. \citet{martone_prototypical_2022} bring this idea into the neuro-symbolic setting through logic tensor networks, and \citet{andolfi_right_2025} combine prototype anchoring with a semantic loss, showing that shortcuts are avoided under a separability condition on the prototypes. These methods supply the perceptual signal that pure soft grounding lacks, but each still compiles the background knowledge into a differentiable, task-specific loss that must be re-derived for every task. Soft-PNet keeps the prototype signal but moves the symbolic constraints out of the loss and into cache membership, leaving a single KL objective that is identical across tasks.

\paragraph{Differentiable neuro-symbolic losses.} A broad family of methods makes reasoning differentiable by relaxing logical semantics: probabilistic inference \citep{manhaeve_deepproblog_2018, choi_probabilistic_2020}, fuzzy logic \citep{badreddine_logic_2022}, knowledge compilation into a semantic loss \citep{ahmed_semantic_2022}, constraint solving as a network layer \citep{wang_satnet_2019}, abductive inference \citep{zhou_abductive_2019}, and program execution over learned concepts \citep{mao_neuro-symbolic_2019, li_closed_2020}. They differ in how constraints are encoded but share a loss that is task-specific and grows with the symbolic complexity of the task. Soft-PNet instead evaluates a fixed objective whose cost does not scale with the number of symbolic operations, because the constraints are precompiled into the cache.

\paragraph{Concept bottleneck models.} Concept Bottleneck Models \citep{koh_concept_2020} and their extensions \citep{espinosa_zarlenga_concept_2022, debot_interpretable_2024} interpose an explicit, supervised concept layer. They assume a fixed concept vocabulary and dense concept annotation, and remain prone to concept leakage when supervision is scarce. Soft-PNet targets the opposite regime, a single anchor per concept with labels given only at the task level, and recovers concepts through the interaction of the anchor signal with the feasible-set cache rather than through direct concept supervision.

\section{Softened Symbol Grounding}\label{apd:background}

Grounding amounts to selecting, for each pair $(\mathbf{x},\mathbf{y})$, the element of $\mathcal{S}(\mathbf{x},\mathbf{y})$ that matches the true concepts. Searching for a deterministic feasible mapping $g$ that maximizes $p_\theta(g(\mathbf{x})\mid\mathbf{x})$ subject to $g(\mathbf{x})\in\mathcal{S}(\mathbf{x},\mathbf{y})$ is fragile, since $g$ is discrete and the network overfits whichever mapping it is given. \citet{li_softened_2024} relax this into a tempered Boltzmann distribution $\tilde{p}_*(\mathbf{c}) \propto p_\theta(\mathbf{c}\mid\mathbf{x})^{1/\gamma}$ over $\mathcal{S}(\mathbf{x},\mathbf{y})$, the closed-form minimizer of an entropy-regularized objective, and sample it with a projection-based Metropolis walk whose acceptance ratio for a move $\mathbf{c}\to\mathbf{c}'$ is $\tau = (p_\theta(\mathbf{c}'\mid\mathbf{x})/p_\theta(\mathbf{c}\mid\mathbf{x}))^{1/\gamma}$, reconstructing feasible states with a satisfiability solver \citep{de_moura_z3_2008}. A single candidate grounding is maintained per data point and updated by this rule.

Concretely, $\tilde{p}_*$ is the distribution $q$ supported on $\mathcal{S}(\mathbf{x},\mathbf{y})$ that maximizes $\mathbb{E}_{\mathbf{c}\sim q}[\log p_\theta(\mathbf{c}\mid\mathbf{x})] + \gamma\, H(q)$, where $H$ is the Shannon entropy: as $\gamma \to 0$ it concentrates on the highest-scoring feasible assignment, recovering a deterministic grounding, and as $\gamma$ grows it approaches the uniform distribution on $\mathcal{S}(\mathbf{x},\mathbf{y})$. Soft-PNet samples this same distribution with two changes: $p_\theta$ is the prototype distribution rather than the classifier output, and a $K$-candidate working set replaces the single grounding. Its soft-target weights $w^{(j)} \propto p_\theta(\mathbf{c}^{(j)}\mid\mathbf{x})^{1/\gamma}$ are exactly $\tilde{p}_*$ restricted to $\mathcal{R}_K$ and renormalized, so the per-concept target $\tilde{p}^i$ is the marginal of this tempered Boltzmann over the cached candidates, and the KL loss (\equationref{eq:kl_loss}) trains the network toward the distribution the walk samples.

\section{Projection Operators}\label{apd:projection}

The Metropolis update in \algorithmref{alg:metro_cache} requires a projection operator $\Pi$ that maps a feasible state to a lower-dimensional space, applies a perturbation, and recovers a new feasible state via the inverse projection $\Pi^{-1}$. We describe the concrete projection strategy used for each task.

\subsection{MNIST-EvenOdd}

Each feasible state is a digit pair $\mathbf{c} = (d_1, d_2)$ with $d_1 + d_2 = n$ and $0 \leq d_1, d_2 \leq 9$. Since the sum fixes the second digit given the first, the projection reduces the pair to its sum, $\Pi(d_1, d_2) = n$. The inverse projection samples uniformly at random a different feasible pair with the same sum from the cache $\mathcal{R}$, and retains the current pair if no alternative exists. The feasible pairs for each sum $n \in \{0, \dots, 18\}$ are enumerated and stored in $\mathcal{R}$ before training, so the inverse projection is a direct lookup and needs no solver call.

\subsection{Visual Sudoku Classification}

Each feasible state is a valid $4 \times 4$ Sudoku board $\mathbf{c} \in \{1,2,3,4\}^{4 \times 4}$ satisfying the standard uniqueness constraint in every row, column, and $2 \times 2$ sub-block. The training set contains both valid (SAT) and invalid (UNSAT) boards, requiring the method to handle both manifolds during training. We formalize this distinction through a corruption operator and a symmetry-based projection, described in turn below.

\paragraph{Corruption operator.}
To generate UNSAT boards from SAT ones in a controlled manner, we define a corruption operator
\begin{align}
    \phi : \mathcal{S}_{\mathrm{SAT}} \rightarrow \mathcal{S}_{\mathrm{UNSAT}},\notag
\end{align}
where $\mathcal{S}_{\mathrm{SAT}}$ denotes the set of all valid $4 \times 4$ Sudoku configurations and $\mathcal{S}_{\mathrm{UNSAT}}$ denotes configurations that violate at least one uniqueness constraint. Concretely, $\phi$ selects uniformly at random a substructure $r \in \{\text{row}, \text{column}, 2\times 2 \text{ sub-block}\}$ of a valid board$\mathbf{c} \in \mathcal{S}_{\mathrm{SAT}}$, then replaces one digit in $r$ with a digit already present in $r$, introducing a repeated value while preserving the overall grid format. By construction, $\phi(\mathbf{c}) \notin \mathcal{S}_{\mathrm{SAT}}$ for all$\mathbf{c} \in \mathcal{S}_{\mathrm{SAT}}$.

The Metropolis walk operates exclusively on $\mathcal{S}_{\mathrm{SAT}}$, ensuring that candidate states always satisfy the Sudoku constraints. UNSAT boards are generated on demand at loss computation time by applying $\phi$ to the current cache state, so the chain itself never leaves the feasible manifold.

\paragraph{Projection operator.} The projection keeps the eight cells of the two diagonal $2 \times 2$ sub-blocks and discards the other eight, $\Pi(\mathbf{c}) = \mathbf{c}|_{\mathcal{K}}$, where $\mathcal{K}$ is the set of kept cells. At each step the values of two cells within one kept sub-block are swapped, using one of ten predefined within-block swaps chosen uniformly at random. The inverse projection looks up a valid board in the cache $\mathcal{R}$ whose kept cells match the perturbed values, and retains the current board if no such board exists. Every reconstructed state is drawn from $\mathcal{R}$, the set of all valid boards, so the walk stays on $\mathcal{S}_{\mathrm{SAT}}$ and moves among distinct feasible boards rather than a single fixed one. Proposals are accepted or rejected by the ratio in \algorithmref{alg:metro_cache}.
\subsection{Kand-Logic}

Each feasible state is a scene configuration $\mathbf{c} \in \{0,1,2\}^{3 \times 3 \times 2}$, where the first two dimensions index $3$ sub-figures of $3$ objects each, and the last dimension indexes the two attributes (shape $\in \{0,1,2\}$ and color $\in \{0,1,2\}$) of each object. The training set contains both valid (SAT) and invalid (UNSAT) scenes, where validity is determined by a logical pattern defined over the primitive attributes. We denote by $\mathcal{S}_{\mathrm{SAT}}$ the set of all scene configurations that satisfy this pattern, and by $\mathcal{S}_{\mathrm{UNSAT}}$ those that violate it.

\paragraph{Corruption operator.}
To generate UNSAT scenes from SAT ones in a controlled manner, we define a corruption operator
\begin{align}
    \phi : \mathcal{S}_{\mathrm{SAT}} \rightarrow \mathcal{S}_{\mathrm{UNSAT}},\notag
\end{align}
where, given a valid scene $\mathbf{c} \in \mathcal{S}_{\mathrm{SAT}}$, $\phi$ selects a randomly chosen object and replaces one of its attributes, shape or color, with an alternative value drawn uniformly from $\{0,1,2\} \setminus \{a\}$, where $a$ is the current attribute value. Validity of the resulting configuration is verified by the \texttt{kand\_label} oracle, which checks whether the modified scene violates the logical pattern; the proposal is accepted as a corrupted sample only if the oracle confirms $\phi(\mathbf{c}) \in \mathcal{S}_{\mathrm{UNSAT}}$. If the oracle rejects the proposal, a new object and attribute are sampled until a valid corruption is found. By construction, $\phi(\mathbf{c}) \notin \mathcal{S}_{\mathrm{SAT}}$ for all accepted outputs.

As with Visual Sudoku, the Metropolis walk operates exclusively on $\mathcal{S}_{\mathrm{SAT}}$, and $\phi$ is applied only at loss computation time to generate the UNSAT supervision signal from the current cache state.

\paragraph{Projection operator.}
For SAT scenes, the projection $\Pi$ operates by applying a random attribute switch to a selected object. Concretely, at each walk step, a randomly chosen object has one of its attributes (shape or color) replaced by a value sampled uniformly from $\{0,1,2\} \setminus \{a\}$, where $a$ is the current attribute value. One object position is held fixed across steps to preserve partial structure during exploration. The inverse projection $\Pi^{-1}$ then verifies whether the resulting scene $\mathbf{c}'$ satisfies the logical pattern via the \texttt{kand\_label} oracle:
\begin{align}
    \Pi^{-1}(u') =
    \begin{cases}
        \mathbf{c}' & \text{if } \mathbf{c}' \in \mathcal{S}_{\mathrm{SAT}}, \\
        \mathbf{c}  & \text{otherwise}, \notag
    \end{cases}
\end{align}
so that infeasible proposals are discarded, and the current state is retained. A fixed number of $n_{\mathrm{steps}} = 1$ attribute switches is applied per Metropolis iteration, and the proposed state is subsequently accepted or rejected according to the acceptance ratio in \algorithmref{alg:metro_cache}.

\section{Experimental Setup}\label{apd:setup}

\subsection{Metrics and Compute}

Concept accuracy is the fraction of individual concepts classified correctly (digits in \texttt{MNIST-EvenOdd} and Visual Sudoku, and shape and color primitives in \texttt{Kand-Logic}), and concept F1 is the macro-averaged F1 over concept classes. Label accuracy and F1 are computed analogously on the final task label, namely the digit sum, board validity, or scene class. Reported training time is wall-clock time per epoch on a single NVIDIA H100 80GB graphics processing unit (GPU). At each encounter of a training pair, the cache performs one projection-based Metropolis proposal per candidate (\algorithmref{alg:metro_cache}), and the KL loss is activated only after the $E_w$ warmup epochs.

\subsection{Baselines}

We compare against three baselines. \textbf{SoftG} \citep{li_softened_2024} keeps a single candidate grounding per example, refines it with the projection-based Metropolis walk, scores it by the network's own output, and trains the network toward that grounding with a cross-entropy loss. \textbf{SoftG-K} extends SoftG to a population of $K$ candidates refined by the same walk and scored by the network output, but reduces the population to its single highest-scoring grounding as the cross-entropy target; it isolates the effect of a candidate population without the prototype signal. \textbf{PNet+SL} \citep{andolfi_right_2025} pairs the same prototypical encoder as Soft-PNet with a task-specific semantic loss compiled from the background knowledge; it supplies the perceptual anchor but requires a differentiable loss re-derived for each task. Soft-PNet differs from SoftG-K in scoring the cache with the prototype distribution and training against the soft per-concept KL target rather than a single hard grounding, and from PNet+SL in encoding the constraints through cache membership rather than a hand-crafted per-task loss.

\subsection{MNIST-EvenOdd}

\subsubsection{Dataset Construction}

The \texttt{MNIST-EvenOdd} task is a constrained variant of MNIST-Addition in which training pairs are restricted to same-parity digit combinations. Concretely, the training support consists of 16 fixed digit pairs: 8 even-even pairs $\{(0,6),(2,8),(4,6),(4,8)\}$ and their reverses, and 8 odd-odd pairs $\{(1,5),(3,7),(1,9),(3,9)\}$ and their reverses. The label for each pair is the digit sum. The training set contains $6{,}720$ samples drawn from these 16 pairs, with images sampled uniformly from the MNIST training split. The test set contains $960$ out-of-distribution mixed-parity pairs drawn exclusively from even-odd and odd-even combinations that do not appear in the training support, sampled from the MNIST test split. This construction ensures that generalization requires digit recognition rather than exploiting parity-level statistics.

\subsubsection{Anchor Set}

A single labeled anchor image per digit class is extracted from the MNIST training split. The anchor set therefore contains exactly 10 images, one per digit $\{0, \dots, 9\}$, selected as the first occurrence of each class in the training data. During training, the anchor set is augmented with $3$ additional copies per image generated via random affine transformations (rotation up to $15^\circ$, translation up to $10\%$, scale $[0.9, 1.1]$, shear up to $10^\circ$), yielding an effective anchor set of $40$ images. Prototype vectors $\bar{\mathbf{z}}_c$ are recomputed at each training step as centroids of the augmented anchor embeddings.

\subsubsection{Neural Architecture}

The encoder $f_\theta$ is a four-block convolutional network. Each block consists of a $3 \times 3$ convolution with padding $1$, batch normalization, rectified linear unit (ReLU) activation, and $2 \times 2$ max-pooling. The hidden and embedding dimensions are both set to $64$, producing a latent representation $\mathbf{z} \in \mathbb{R}^{64}$. This architecture follows the standard prototypical network backbone \citep{martone_prototypical_2022, andolfi_right_2025}.

For the SoftG and SoftG-K baselines, the encoder is replaced by a LeNet architecture consisting of two convolutional layers with channel sizes $(1, 6, 16)$ and kernel size $5$, each followed by ReLU and $2 \times 2$ max-pooling, and three fully connected layers of sizes $256 \rightarrow 120 \rightarrow 84 \rightarrow 10$.

\subsubsection{Training Details}

All models are trained for $10$ epochs using the Adam optimizer with learning rate $10^{-3}$. The batch size is $64$. For Soft-PNet, the episodic prototypical loss is computed over $50$ episodes per minibatch 
step with $k_{\text{support}} = 1$. The KL loss weight is fixed at $\lambda = 5.0$, and the temperature $\gamma$ decays exponentially ($\gamma \leftarrow \max(0.01,\, 0.95\,\gamma)$ per epoch). On this task the cache target is the single highest-scoring candidate, so \equationref{eq:kl_loss} is applied in its zero-temperature limit, a cross-entropy toward that candidate; Visual Sudoku and \texttt{Kand-Logic} instead use the temperature-weighted soft target. The cache contains $K = 20$ candidates per training sample, initialized by drawing a larger pool from the global solution cache $\mathcal{R}$ and retaining the top-$K$ by prototype score; it is refreshed by one Metropolis step per epoch. 

The global solution cache $\mathcal{R}$ is precomputed by enumerating all valid digit pairs for each possible sum $n \in \{0, \dots, 18\}$, yielding $100$ ordered pairs in total. The projection strategy is described in \appendixref{apd:projection}.

Reproducibility is ensured by running ten independent seeds from the collection $\{i \cdot 128\}_{i=0}^{9}$. All results report the mean and standard deviation over these ten runs.

\subsection{Visual Sudoku Classification}

\begin{figure}[htbp]
    \centering
    \includegraphics[width=\linewidth]{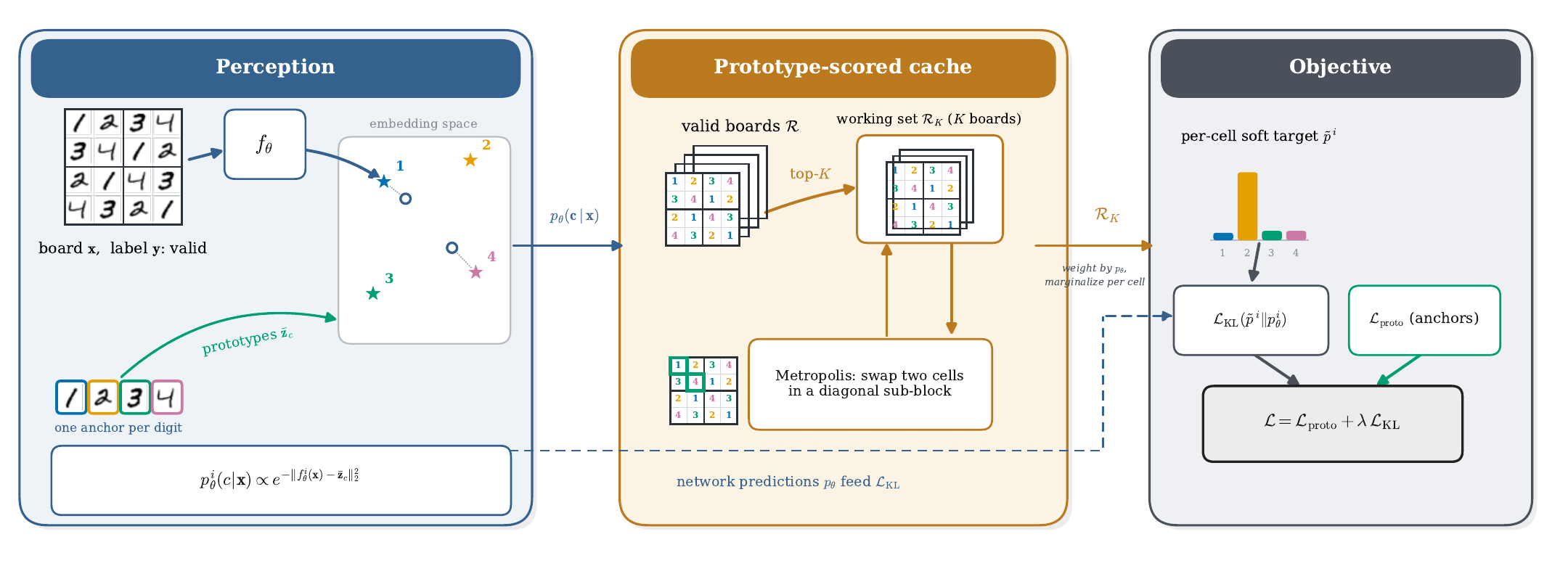}
    \caption{The Soft-PNet pipeline instantiated on Visual Sudoku. The encoder embeds each board cell, and a single labeled anchor per digit builds the prototypes that define $p_\theta$. The precomputed cache of valid boards is scored by the prototype distribution and refined by a projection-based Metropolis walk that swaps two cells within a diagonal sub-block; the per-cell marginals of the prototype-weighted working set form the soft targets of the KL loss, combined with the prototypical loss on the anchors. This mirrors \figureref{fig:pipeline} for the Visual Sudoku task.}
    \label{fig:pipeline_sudoku}
\end{figure}

\figureref{fig:pipeline_sudoku} illustrates the complete Soft-PNet pipeline instantiated for Visual Sudoku.

\subsubsection{Dataset Construction}

Sudoku boards are $4 \times 4$ grids populated with digits from $\{1, 2, 3, 4\}$ subject to the standard uniqueness constraint: each digit appears exactly once in every row, column, and $2 \times 2$ sub-block. The dataset contains equal numbers of valid (SAT) and invalid (UNSAT) boards. Valid boards are generated by sampling from the manifold of feasible $4 \times 4$ Sudoku configurations using a tensorized sampler. Invalid boards are generated by introducing a repeated digit into a randomly selected row, column, or sub-block of an otherwise valid board, violating the uniqueness constraint while preserving the grid format. Each cell of a board is populated with a randomly selected MNIST image of the corresponding digit, drawn independently from the MNIST training or test split depending on whether the board belongs to the training or test set. The training set contains $300$ boards ($150$ SAT and $150$ UNSAT), with the test set fixed at $1{,}000$ boards.

\subsubsection{Anchor Set}

A single labeled anchor image per digit class is sampled from the MNIST training split for digits $\{1, 2, 3, 4\}$. The anchor set therefore contains exactly 4 images. During training, each anchor image is augmented with $9$ additional copies generated via random affine transformations (rotation up to $15^\circ$, translation up to $10\%$, scale $[0.9, 1.1]$, shear up to $10^\circ$), yielding an effective anchor set of $40$ images. Prototype vectors $\bar{\mathbf{z}}_c$ are recomputed at each training step as centroids of the augmented anchor embeddings. A separate anchor image is also used for the Metropolis walk, sampled independently from the training split to avoid overlap with the episodic support set.

\subsubsection{Neural Architecture}

The encoder $f_\theta$ is a four-block convolutional network identical to the one used for \texttt{MNIST-EvenOdd}: each block consists of a $3 \times 3$ convolution with padding $1$, batch normalization, ReLU activation, and $2 \times 2$ max-pooling, with hidden and embedding dimensions set to $64$, producing $\mathbf{z} \in \mathbb{R}^{64}$. The encoder is applied independently to each of the $16$ cells of the board, producing a tensor of cell embeddings of shape $(B, 16, 64)$. The concept distribution $p_\theta^i(c \mid \mathbf{x})$ is computed for each cell via softmax over negative squared distances to the four digit prototypes.

For the SoftG and SoftG-K baselines, the per-cell encoder is replaced by a LeNet with two convolutional layers of channel sizes $(1, 6, 16)$ and kernel size $5$, followed by ReLU and $2 \times 2$ max-pooling, and three fully connected layers of sizes $256 \rightarrow 120 \rightarrow 84 \rightarrow 4$.

\subsubsection{Training Details}

All models are trained for $20$ epochs following the training regime of \citet{li_softened_2024}. The Adam optimizer is used with learning rate $10^{-3}$ and batch size $64$. For Soft-PNet, the episodic prototypical loss is computed over $50$ episodes per minibatch step with $k_{\text{support}} = 1$. The KL loss weight is fixed at $\lambda = 10.0$, the temperature $\gamma$ decays exponentially, and the cache is refreshed by a Metropolis step every two epochs.

The cache contains $K = 100$ candidates per training sample. The global solution cache $\mathcal{R}$ is precomputed before training by enumerating all $288$ valid $4 \times 4$ Sudoku configurations with the tensorized sampler. At initialization, $M = K \cdot 10$ candidates are drawn from $\mathcal{R}$ and the top-$K$ by prototype score are retained as the working set $\mathcal{R}_K$. The Metropolis walk applies the projection strategy described in \appendixref{apd:projection}, operating exclusively on valid boards. Invalid boards in the training set are generated from valid ones via the corruption operator $\phi$ at loss computation time, ensuring the Metropolis chain always explores the manifold of feasible configurations.

Reproducibility is ensured by running ten independent seeds from the collection $\{i \cdot 128\}_{i=0}^{9}$. All results report the mean and standard deviation over these ten runs.

\subsection{Kand-Logic}

\subsubsection{Dataset Construction}

\begin{wrapfigure}{l}{0.45\textwidth}
    \centering
    \includegraphics[width=0.9\linewidth]{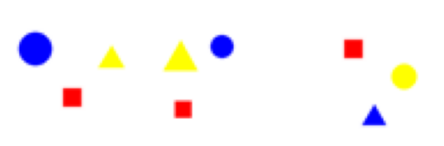}
    \caption{Data point from \texttt{Kand-Logic}.}
    \label{fig:kand_data}
\end{wrapfigure}

The \texttt{Kand-Logic} task is based on the Kandinsky Patterns benchmark \citep{muller_kandinsky_2021}. Each scene consists of $3$ sub-figures, each containing $3$ objects. Every object is defined by two attributes: shape $\in \{\text{square}, \text{circle}, \text{triangle}\}$ and color $\in \{\text{red}, \text{yellow}, \text{blue}\}$, yielding $9$ primitive classes defined by the joint attribute pair (shape, color). Each primitive is represented as a small red-green-blue (RGB) image of the corresponding geometric object. An example is depicted in \figureref{fig:kand_data}. The scene label indicates whether the scene satisfies a logical pattern over the primitive attributes: a scene is valid if, for at least one attribute (shape or color), all three sub-figures share the same repetition type, where a sub-figure's repetition type for that attribute is one of \emph{all three equal}, \emph{all three distinct}, or \emph{exactly two equal}. Formally, writing $t_a(s) \in \{\text{all-equal}, \text{all-distinct}, \text{two-equal}\}$ for the repetition type of the three objects' values of attribute $a$ in sub-figure $s$, the scene is valid if and only if $t_a(1) = t_a(2) = t_a(3)$ for at least one attribute $a \in \{\text{shape}, \text{color}\}$. The dataset contains $3{,}000$ training scenes and a held-out test set, loaded from preprocessed \texttt{.pt} files. The solution space for this task is combinatorially intractable: the number of valid scene configurations grows exponentially with the number of objects and attribute combinations, making exhaustive enumeration infeasible. The global cache $\mathcal{R}$ therefore contains only a representative subset of feasible configurations, sampled before training.

\subsubsection{Anchor Set}

A single labeled anchor image per primitive class is extracted from the training set. The anchor selection procedure iterates over the training scenes and records the first occurrence of each of the $9$ primitive classes, identified by the joint (shape, color) label encoded as $\text{shape} \times 3 + \text{color} \in \{0, \dots, 8\}$. The anchor set therefore contains exactly $9$ images. During training, each anchor image is augmented with $3$ additional copies via random affine transformations (rotation up to $15^\circ$, translation up to $10\%$, scale $[0.9, 1.1]$, shear up to $10^\circ$), yielding an effective anchor set of $36$ images. Prototype vectors are organized as a $(3, 3, D)$ joint grid over the two attributes shape and color, and per-concept marginal distributions over colors and shapes are obtained by summing the joint softmax distribution over the complementary axis.

\subsubsection{Neural Architecture}

The encoder $f_\theta$ is a four-block convolutional network with $3$-channel RGB input and hidden and embedding dimensions set to $64$, producing $\mathbf{z} \in \mathbb{R}^{64}$. Each block consists of a $3 \times 3$ convolution with padding $1$, batch normalization, ReLU activation, and $2 \times 2$ max-pooling. The encoder is applied independently to each of the $9$ object images in a scene (3 sub-figures $\times$ 3 objects), producing a tensor of object embeddings of shape $(B, 9, 64)$. The concept distribution is computed as a softmax over negative squared distances to the $9$ primitive prototypes arranged in the $(3, 3, D)$ grid, and marginals over shape and color are obtained by summing the joint distribution over the complementary axis.

\subsubsection{Training Details}

All models are trained for $20$ epochs using the Adam optimizer with learning rate $10^{-3}$, weight decay $10^{-4}$, and batch size $64$. A step learning rate scheduler reduces the learning rate by a factor of $0.5$ every $3$ epochs. For Soft-PNet, the episodic prototypical loss is computed over $50$ episodes per minibatch step with $k_{\text{support}} = 1$. The KL loss weight is held at zero for a $3$-epoch warmup, then ramped linearly to $\lambda = 1.0$ over the next $5$ epochs. The temperature $\gamma$ decays exponentially with factor $0.95$ per epoch, floored at $0.01$.

The cache contains $K = 20$ candidates per training sample. At initialization, $M = K \cdot 1000$ candidates are drawn from the global cache $\mathcal{R}$ and filtered to retain only those consistent with the observed scene label; then the top-$K$ by prototype score are selected as the working set $\mathcal{R}_K$. The Metropolis walk applies the scene predicate switch projection described in \appendixref{apd:projection}, operating exclusively on feasible scene configurations. Invalid scenes used for label-level supervision are generated from valid ones at loss computation time via the corruption operator $\phi$, which introduces a constraint violation by modifying a randomly selected object attribute.

Reproducibility is ensured by running ten independent seeds from the collection $\{i \cdot 128\}_{i=0}^{9}$. All results report the mean and standard deviation over these ten runs.

\section{Cache Subroutines}\label{apd:training}

Soft-PNet maintains, for each training pair $(\mathbf{x}, \mathbf{y})$, a working set $\mathcal{R}_K$ of $K$ feasible candidate solutions. Two subroutines manage this set: \algorithmref{alg:init_cache} constructs it the first time a pair is encountered, and \algorithmref{alg:metro_cache} refines it on every subsequent encounter. Both are called by the training loop in \algorithmref{alg:softpnet_training}, and both score candidates with the current prototype distribution $p_\theta(\cdot \mid \mathbf{x})$, so the working set tracks the model as training proceeds.

\begin{algorithm2e}[H]
\caption{Cache Initialization}
\label{alg:init_cache}
\KwIn{Pair $(\mathbf{x}, \mathbf{y})$, global cache $\mathcal{R}$, 
      network $p_\theta$, integers $K, N$}
\KwOut{Working set $\mathcal{R}_K$}
Filter $\mathcal{R}$ to obtain $\mathcal{R}(\mathbf{x}, \mathbf{y}) = 
    \{ \mathbf{c} \in \mathcal{R} \mid f_{\mathbf{K}}(\mathbf{c}) = \mathbf{y} \}$\;
Sample $M = K \cdot N$ candidates from $\mathcal{R}(\mathbf{x}, \mathbf{y})$\;
Score each candidate $s^{(j)} = \log p_\theta(\mathbf{c}^{(j)} \mid \mathbf{x})$
    for $j = 1, \dots, M$\;
Select working set $\mathcal{R}_K \leftarrow \mathrm{top}\text{-}K\{s^{(j)}\}$\;
\end{algorithm2e}

\paragraph{Initialization.} The first time a pair $(\mathbf{x}, \mathbf{y})$ is seen, \algorithmref{alg:init_cache} restricts the global cache $\mathcal{R}$ to the solutions feasible for the observed label, $\mathcal{R}(\mathbf{x}, \mathbf{y}) = \{ \mathbf{c} \in \mathcal{R} \mid f_{\mathbf{K}}(\mathbf{c}) = \mathbf{y} \}$, so every candidate satisfies the background knowledge by construction. It then draws an oversampled pool of $M = K \cdot N$ candidates from this set, scores each by its log-probability $\log p_\theta(\mathbf{c} \mid \mathbf{x})$ under the current model, and keeps the $K$ highest-scoring candidates as the working set $\mathcal{R}_K$. Oversampling before selecting the top $K$ yields a working set that is feasible and already consistent with the model, which is a stronger starting point for the subsequent walk than a single random candidate.

\begin{algorithm2e}[H]
\caption{Cache Metropolis Update}
\label{alg:metro_cache}
\KwIn{Input $\mathbf{x}$, working set $\mathcal{R}_K$, global cache $\mathcal{R}$,
      network $p_\theta$, temperature $\gamma$}
\KwOut{Updated working set $\mathcal{R}_K$}
\For{each $\mathbf{c}^{(j)} \in \mathcal{R}_K$}{
    Compute $u^{(j)} = \Pi(\mathbf{c}^{(j)})$, perturb to obtain ${u'}^{(j)}$\;
    Recover ${\mathbf{c}'}^{(j)} = \Pi^{-1}({u'}^{(j)})$ via lookup in $\mathcal{R}$\;
    Compute $\tau^{(j)} = \left( 
        \frac{p_\theta({\mathbf{c}'}^{(j)} \mid \mathbf{x})}
             {p_\theta(\mathbf{c}^{(j)} \mid \mathbf{x})} 
        \right)^{1/\gamma}$\;
    Sample $\nu \sim \mathcal{U}[0,1]$\;
    \If{$\nu \leq \tau^{(j)}$}{
        Replace $\mathbf{c}^{(j)}$ with ${\mathbf{c}'}^{(j)}$ in $\mathcal{R}_K$\;
    }
}
\end{algorithm2e}

\paragraph{Metropolis update.} On subsequent encounters, \algorithmref{alg:metro_cache} advances each candidate in $\mathcal{R}_K$ by one projection step. The candidate is mapped to a lower-dimensional space by the projection operator $\Pi$ (\appendixref{apd:projection}), perturbed there, and lifted back to a feasible neighbor ${\mathbf{c}'}^{(j)}$ by the inverse projection $\Pi^{-1}$, which is a lookup into $\mathcal{R}$ and so always returns a feasible state. The proposal is accepted with the tempered Metropolis ratio $\tau^{(j)} = \big( p_\theta({\mathbf{c}'}^{(j)} \mid \mathbf{x}) / p_\theta(\mathbf{c}^{(j)} \mid \mathbf{x}) \big)^{1/\gamma}$; drawing $\nu \sim \mathcal{U}[0,1]$ and accepting when $\nu \leq \tau^{(j)}$ is equivalent to accepting with probability $\min(1, \tau^{(j)})$, so a candidate always moves to a higher-scoring neighbor and only occasionally to a lower-scoring one. The temperature $\gamma$ controls this exploration: as it is annealed toward zero the walk becomes increasingly greedy and the working set concentrates on the candidates that the prototype distribution scores highest.

\section{Ablation Study}\label{apd:ablation}

We evaluate the individual contribution of the three core components of Soft-PNet on Visual Sudoku Classification: prototype anchor grounding, the $K$-candidate population cache, and the structured projection walk. Each variant removes one component while keeping the other two intact.

\textbf{Abl-A (no anchor).} The prototype distribution is replaced by a uniform distribution over the feasible set, so the Metropolis acceptance ratio and the top-$K$ scoring are driven by the raw network output $p_\theta$ alone, without any geometric reference to the labeled anchors. This is equivalent to the SoftG-K baseline equipped with the prototypical encoder but without anchor-guided scoring.

\textbf{Abl-B (no population).} The cache is collapsed to $K = 1$, meaning a single candidate is maintained per training sample and used as a hard label rather than a soft target distribution. The prototype anchor and projection walk are retained, but the KL loss degenerates to a cross-entropy over a single grounding.

\textbf{Abl-C (no projection).} The structured projection walk is replaced by naive single-cell digit-flip proposals in the unrestricted digit space. Feasibility is checked by the solver, and infeasible proposals are rejected. The prototype anchor and population cache are retained.

Results are reported in \tableref{tab:ablation}. Removing the anchor (Abl-A) produces the largest degradation: board F1 collapses from $88.48 \pm 5.18\%$ to $32.03 \pm 16.34\%$, while digit accuracy drops to $89.97 \pm 3.72\%$. The concept-label gap reopens, confirming that without perceptual grounding the cache diversifies structurally but remains semantically arbitrary. Removing the population (Abl-B) causes a different failure mode: both digit accuracy ($72.99 \pm 10.92\%$) and board accuracy ($50.64 \pm 0.51\%$) degrade substantially, showing that a single candidate grounding is fragile even when the anchor provides a correct perceptual signal, because one sample is insufficient to estimate the soft target distribution reliably. Removing the projection walk (Abl-C) has only a marginal effect on performance ($98.46\%$ digit accuracy, $87.25\%$ board F1), indicating that the prototype-guided scoring is the dominant driver of cache quality, with the projection contributing primarily to exploration diversity across connected regions of the feasible manifold.

\section{Qualitative Concept Recovery}\label{apd:concept}

\begin{figure}[htbp]
    \centering
    \includegraphics[width=\linewidth]{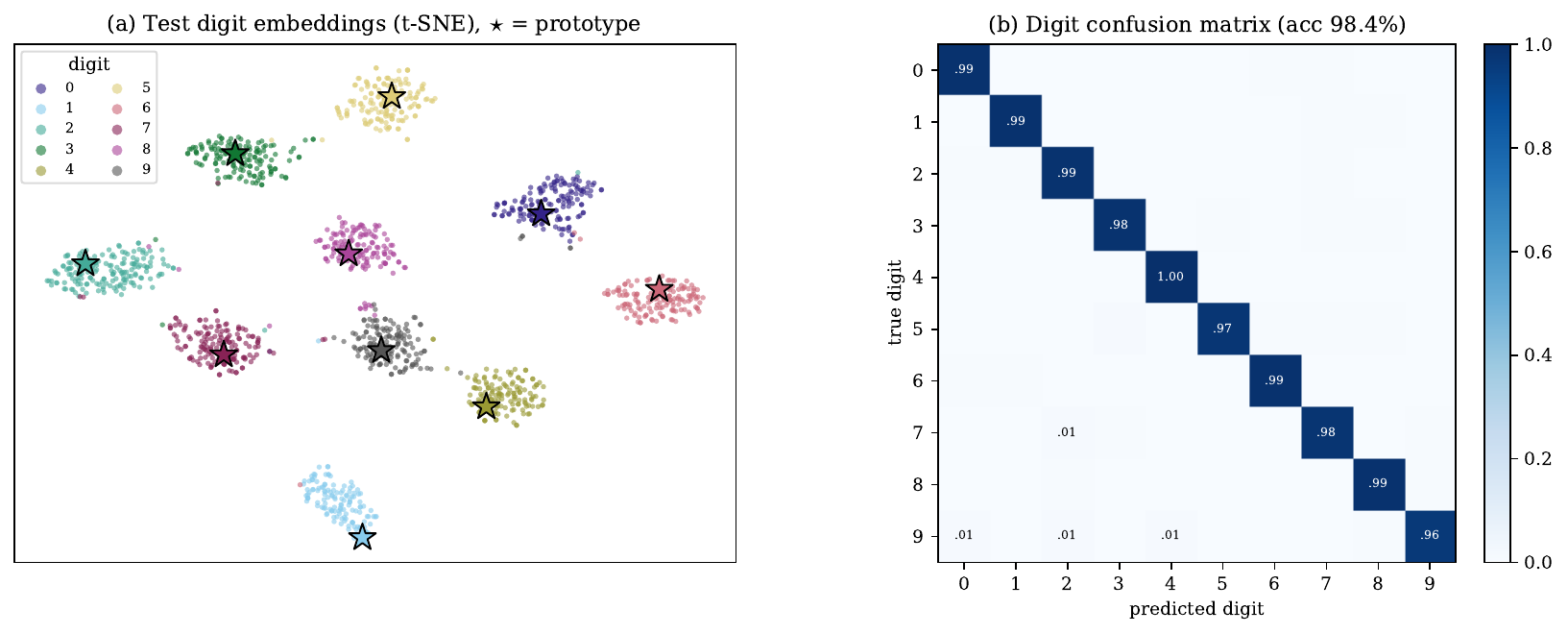}
    \caption{Concept recovery on the out-of-distribution \texttt{MNIST-EvenOdd} test set for a representative Soft-PNet run. (a) t-SNE of the test digit embeddings colored by true digit, with each class prototype marked as a star. (b) Digit confusion matrix. Both indicate that the network recovers the true digit concepts under label-only supervision.}
    \label{fig:concept_recovery}
\end{figure}

\figureref{fig:concept_recovery} visualizes the concepts that Soft-PNet recovers on the \texttt{MNIST-EvenOdd} test set, which contains only mixed-parity digit pairs that never appear during training. Panel (a) projects the test digit embeddings to two dimensions with t-SNE, colored by ground-truth digit, and marks the learned prototype of each class with a star: the ten digits form well-separated clusters centered on their prototypes, so the embedding generalizes to the unseen parity combinations rather than collapsing onto a label-satisfying shortcut. Panel (b) reports the digit confusion matrix, whose strong diagonal confirms that the recovered concepts match the true digits.

\end{document}